\documentclass[conference, letter]{IEEEtran}
\usepackage{times}
\usepackage{graphicx}
\usepackage{bm}
\usepackage{amsmath}
\usepackage{amsfonts}
\usepackage{subcaption}

\usepackage[numbers]{natbib}
\usepackage{multicol}
\usepackage[bookmarks=true]{hyperref}
\usepackage[T1]{fontenc}

\begin{document}


\title{BayesSim: adaptive domain randomization via probabilistic inference for robotics simulators}


\author{\authorblockN{
Fabio Ramos\authorrefmark{1}\authorrefmark{2}
Rafael Carvalhaes Possas\authorrefmark{1}\authorrefmark{2}
Dieter Fox\authorrefmark{1}\authorrefmark{3}
\\
\authorblockA{\authorrefmark{1}NVIDIA \hspace{5mm}
\authorrefmark{2}University of Sydney \hspace{5mm}
\authorrefmark{3}University of Washington}
}}



%

\maketitle
\vspace{-0.5cm}
\begin{abstract}
We introduce BayesSim~\footnote{Code available at \url{https://github.com/rafaelpossas/bayes_sim}}, a framework for robotics simulations allowing a full Bayesian treatment for the parameters of the simulator. As simulators become more sophisticated and able to represent the dynamics more accurately, fundamental problems in robotics such as motion planning and perception can be solved in simulation and solutions transferred to the physical robot. However, even the most complex simulator might still not be able to represent reality in all its details either due to inaccurate parametrization or simplistic assumptions in the dynamic models.  
BayesSim provides a principled framework to reason about the uncertainty of simulation parameters. Given a black box simulator (or generative model) that outputs trajectories of state and action pairs from unknown simulation parameters, followed by trajectories obtained with a physical robot, we develop a likelihood-free inference method that computes the posterior distribution of simulation parameters. This posterior can then be used in problems where Sim2Real is critical, for example in policy search. We compare the performance of BayesSim in obtaining accurate posteriors in a number of classical control and robotics problems. Results show that the posterior computed from BayesSim can be used for domain randomization outperforming alternative methods that randomize based on uniform priors. 
\end{abstract}

\IEEEpeerreviewmaketitle

\section{Introduction}
Simulators are emerging as one of the most important tools for efficient learning in robotics. With physically accurate and photo-realistic simulation, perception models and control policies can be trained more easily before being transferred to real robots, saving both time and costs of running complex experiments. Unfortunately, in many cases, models and policies trained in simulation are not seamlessly transferable to the real systems. Lack of knowledge about the correct simulation parameters, oversimplified simulation models, or insufficient numerical precision for differential equation solvers can all play a significant role in this problem.  To ameliorate this problem, a popular approach is to \emph{sample} different simulation parameters during training and thereby learn models that are robust to simulation perturbations. This approach, often referred to as domain randomization (DR), has been shown to perform surprisingly well in areas such as learning to control a humanoid robot~\cite{Mor15Ens}, manipulate table top objects~\cite{domrandom}, estimating 6D object poses from images~\cite{Tre18Dee}, or dexterous in-hand manipulation~\cite{andrychowicz2018learning}. 

A crucial question regarding domain randomization is which simulation parameters to randomize over and from which distributions to sample their values from. Typically, these parameters and their distributions are determined in a manual process by iteratively testing whether a model learned in randomized simulation works well on the real system.  If the model does not work on the real robot, the randomization parameters are changed so that they better cover the conditions observed in the real world. To overcome this manual tuning process,
 \cite{Che19Sim} recently showed how policy executions on a real robot can be used to automatically update a Gaussian distribution over the sampling parameters such that the simulator better matches reality. However, by restricting sampling distributions to Gaussians, this approach cannot model more complex uncertainties and dependencies among parameters.  Alternatively, one could perform system identification to better estimate simulation parameters from the real data.  Since most of these techniques assume that the simulation equations are known and only provide point estimates for the parameters, they do not account for the uncertainties associated with the measurement process, numerical precision of differential equation solvers, or simplistic models~\cite{goodwin1977systemid}. 

In this paper we provide a principled Bayesian method to compute full posteriors over simulator parameters, thereby overcoming the limitations of previous approaches.  Our technique, called BayesSim, leverages recent advances in likelihood-free inference for Bayesian analysis to update posteriors over simulation parameters based on small sets of observations obtained on the real system. The main difficulty in computing such posteriors relates to the evaluation of the likelihood function, which models the relationship between simulation parameters and corresponding system behavior, or observations in the real world.  While a simulator \emph{implicitly} defines this relationship, the likelihood function requires the \emph{inverse} of the simulator model, \emph{i.e.}, how observed system behavior can be used to derive corresponding simulation parameters.  Importantly, BayesSim does not assume access to the internal differential equations underlying the simulator and treats the simulator as a black box. 

We make the following contributions: First, we introduce BayesSim as a generic framework for probabilistic inference with robotics simulators and show that it can provide a full space of simulation parameters that best fit observed data. This is in contrast to traditional system identification methods that only provide the best fitting solution. Second, we propose a novel mixture density random Fourier network to approximate the conditional distribution $p(\bm{\theta}|\mathbf{x}^r)$ directly by learning from pairs $\{\bm{\theta}_i,\mathbf{x}_i^s\}_{i=1}^N$ generated from the proposal prior and the simulator. Finally, we show that learning policies with domain randomization where the simulator parameters are randomized according to the posterior provided by BayesSim generates policies that are significantly more robust and easier to train than randomization directly from the prior. 

\section{Related Work}

Simulators accelerate machine learning impact by allowing faster, highly-scalable and low cost data collection. Many other scientific domains such as economics \cite{gourieroux1993indirect}, evolutionary biology \cite{beaumont2002approximate} and cosmology \cite{schafer2012likelihood} also rely on simulator-based modelling to provide further advancements in research.  In robotics, "reality gap" is not only seen in control, robotics vision is also affected by this problem \cite{domrandom}. Algorithms trained on images from a simulation can frequently fail on different environments as the appearance of the world can differ greatly from one system to the other.

Randomizing the dynamics of a simulator while training a control policy has proven to mitigate the reality gap problem \cite{robcontrol}. Simulation parameters could vary from physical settings like damping, friction and object masses \cite{robcontrol} to visual parameters like objects textures, shapes and etc \cite{domrandom}. Another similar approach is that of adding noise to the system parameters \cite{quadrupeds} instead of sampling new parameters from a uniform prior distribution. Perturbation can also be seen on robot locomotion \cite{mordatch2015ensemble} where planning is done through an ensemble of perturbed models. Lastly, interleaving policy roll outs between simulation and reality has also proven to work well on swing-peg-in-hole and opening a cabinet drawer tasks \cite{chebotar2018closing}.

Learning models from simulations of data can leverage one's understanding of the physical world potentially helping to solve the aformentioned problem. Until recently, Approximate Bayesian Computation \cite{beaumont2002approximate} has been one of the main methods used to tackle this type of problem. Rejection ABC \cite{pritchard1999population} is the most basic method where parameter settings are accepted/rejected if they are within a certain specified range. The set of accepted parameters approximates the posterior for the real parameters. Markov Chain Monte Carlo ABC (MCMC-ABC) \cite{marjoram2003markov} improves over its precedent by perturbing accepted parameters instead of independently proposing new parameters. Lastly, Sequential Monte Carlo ABC (SMC-ABC) \cite{bonassi2015sequential} leverages sequential importance sampling to simulate slowly-change distributions where the last one is an approximation of the true parameter posterior. In this work, we use a $\epsilon$-free approach \cite{epsilonfree} for likelihood-free inference, where a Mixture of Density Random Fourier Network estimates the parameters of the true posterior through a Gaussian mixture.

\begin{figure}
\centering

\includegraphics[width=0.40\textwidth]{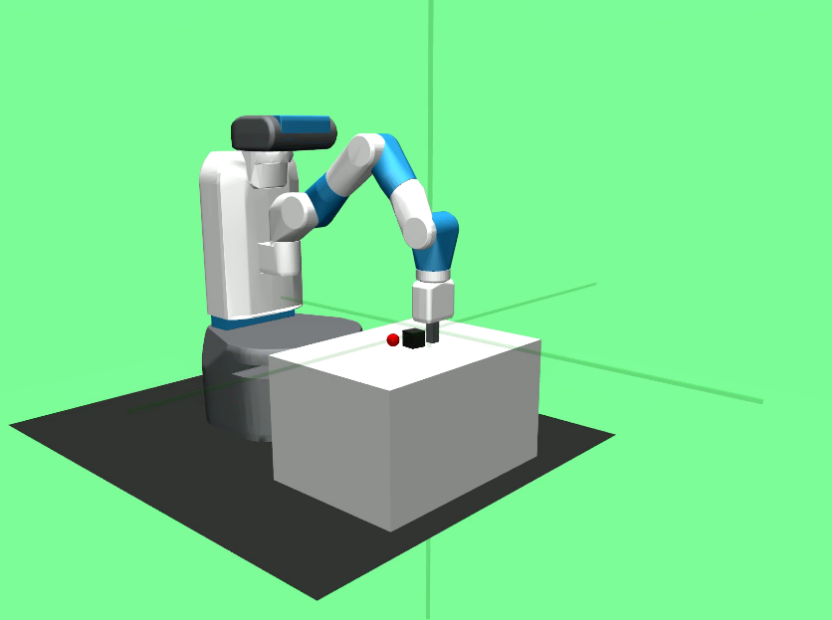}

\caption[Fetch Robot Tasks]{Fetch Push and Sliding tasks: the robot has full access to the entire table and multiple iterations with the object (pushing) or one shot at pushing the object to its target (sliding).} 
\label{fig:fetch_push}
\end{figure}

A wide range of complex robotics control problems have been recently solved using Deep Reinforcement Learning (Deep RL) techniques \cite{andrychowicz2018learning,robcontrol, quadrupeds}. Classic control problems like Pendulum, Mountain Car, Acrobot and Cartpole have been successfully tackled using policy search with algorithms like Trust Region Policy Optimization (TRPO) \cite{schulman2015trust} and Proximal Policy Optimization (PPO) \cite{schulman2017proximal}. More complex tasks in robotics such as the ones in manipulation are still difficult to solve using traditional policy search. Both Push and Slide tasks (Figure \ref{fig:fetch_push}) on the fetch robot \cite{brockman2016openai} were only solved recently using the combination of Deep Deterministic Policy Gradients (DDPG) \cite{lillicrap2015continuous} and Hindsight Experience Replay (HER) \cite{hindsight2017}.

 \section{Preliminaries}
 In this section we provide background on likelihood free inference and reinforcement learning. As we shall see, policy search via domain randomization is one of the applications in which BayesSim proved to be valuable. 
 
 \subsection{Likelihood-free inference}
  
 BayesSim takes a \emph{prior} $p(\bm{\theta})$ over simulation parameters $\bm{\theta}$, a black box generative model or simulator $\mathbf{x}^{s} = g(\bm{\theta})$ that generates simulated observations $\mathbf{x}^{s}$ from these parameters, and observations from the physical world $\mathbf{x}^{r}$ to compute the posterior  $p(\bm{\theta}|\mathbf{x}^{s},\mathbf{x}^{r})$.
The main difficulty in computing this posterior relates to the evaluation of the likelihood function $p(\mathbf{x}|\bm{\theta})$ which is defined \emph{implicitly} from 
the simulator~\cite{implicit84}. Here we assume that the simulator is a set of dynamical differential equations associated with a numerical or analytical solver which are typically intractable and expensive to evaluate. Furthermore, we do not assume these equations are known and treat the simulator as a black box. This allows our method to be utilized with many robotics simulators (even closed source ones) but requires a method where the likelihood cannot be evaluated directly but instead only sampled from, by performing forward simulations. This is referred to in statistics as likelihood-free inference of which the most popular family of algorithms to address it are known as approximate Bayesian computation (ABC)~\cite{beaumont2002approximate, marjoram2003markov, sequentialABC07}. 

In ABC, the simulator is used to generate synthetic observations from samples following the parameters prior. These samples are accepted when features or sufficient statistics computed from the synthetic data are similar to those from real observations obtained from physical experiments. As a sampling-based technique, ABC can be notoriously slow to converge, particularly when the dimensionality of the parameter space is large. Formally,  
 ABC approximates the posterior $p(\bm{\theta}|\mathbf{x}=\mathbf{x}^r) \propto p(\mathbf{x}=\mathbf{x}^r|\bm{\theta})p(\bm{\theta})$ using the Bayes' rule. However as the likelihood function $p(\mathbf{x}=\mathbf{x}^r|\bm{\theta})$ is not available, conventional methods for Bayesian inference cannot be applied. ABC sidesteps this problem by approximating $p(\mathbf{x}=\mathbf{x}^r|\bm{\theta})$ by $p(\parallel\mathbf{x}-\mathbf{x}^r\parallel<\epsilon|\bm{\theta})$, where $\epsilon$ is a small value defining a sphere around real observations $\mathbf{x}^r$, and using Monte Carlo to estimate its value. The quality of the approximation increases as $\epsilon$ decreases however, the computational cost can become prohibitive as most simulations will not fall within the acceptable region. 
  
 \subsection{Reinforcement learning and policy search in robotics}
 
 We consider the default RL scenario where an agent interacts in discrete timesteps with an environment $\mathbf{E}$. At each step $t$ the agent receives an observation $\mathbf{o}_t$, takes an action $\mathbf{a}_t$ and receives a real number reward $r_t$. In general, actions in robotics are real valued $\mathbf{a}_t\in \mathbb{R}^D$ and environments are usually partially observed so that the entire history of observation, action pairs $\bm\eta  = \{\mathbf{s}_t,\mathbf{a}_t,\mathbf{o}_t\}_{t=0}^{T-1}$. The goal is to maximize the expected sum of discounted future rewards by following a policy $\pi(\mathbf{a}_t|\mathbf{s}_t;\bm\beta)$, parametrized by $\bm\beta$,
  \begin{equation}
  J(\bm\beta) = \mathbb{E}_{\bm\eta}\left[ \sum_{t=0}^{T-1} \gamma^{(t)}r(\mathbf{s}_t, \mathbf{a}_t) | \bm\beta \right]. 
  \label{eq:cost}
 \end{equation}
 
  Many approaches in reinforcement learning make use of the recursive relationship known as the Bellman equation where $Q^\pi$ is the action-value function describing the expected return after taking an action $\mathbf{a}_t$, in state $\mathbf{s}_t$ and thereafter following policy $\pi$.
 \small
 \begin{equation}
{Q^\pi(\mathbf{s}_t, \mathbf{a}_t)} =  {\mathbb{E}_{r_t, s_{t+1}}} {[r(\mathbf{s}_t,\mathbf{a}_t)+\gamma\mathbb{E}_{a_{t+1}}[Q^\pi(\mathbf{s}_{t+1},\mathbf{a}_{t+1})]]}
\end{equation}
\normalsize
In recent years, the advancements in traditional RL methods have allowed their application to control tasks with continuous action spaces. Inheriting ideas from DQN \cite{mnih2015human}, Deep Deterministic Policy Gradients have been relatively successful in a wide range of control problems. The main caveat of DDPG algorithms is that they rely on efficient experience sampling to perform well. Improving the way how experience is collected is one of most important topics in today's RL community. Experience Replay \cite{lin1992self} and Prioritized Experience replay \cite{prioritized} still performs poorly in a repertoire of robotics tasks where the reward signal is sparse. Hindsight Experience replay (HER) \cite{hindsight2017}, on the other hand, performs well in this scenario as it  breaks down single trajectories/goals into smaller ones and, thus, provides the policy optimization algorithm with better reward signals. HER has been mostly based in a recent RL concept: Multi-Goal learning with Universal Function Approximators \cite{schaul2015universal}.

Another set of successful policy search algorithms is based on optimization through trust regions. They are less sensitive to the experience sampling problem mentioned above. The maximum step size for exploration is determined by its trust region and the optimal point is then evaluated progressively until convergence has been reached. The main idea is that updates are always limited by their own trust region, and, therefore, learning speed is better controlled. Proximal Policy Optimization \cite{schulman2017proximal} and Trust Region Policy optimization \cite{schulman2015trust} have applied these ideas providing state of the art performance in a wide range of control problems.

Both techniques differ on the way they sample experiences. While the first is an off-policy algorithm - experiences are generated by a behaviour policy, the second is an on-policy algorithm where the policy used to generated experience is the same used to perform the control task. These algorithms will have comparable performance on different robotics control scenarios therefore should be considered the current state of the art on such problems.

\section{BayesSim}
\label{sec:BayesSim}

\subsection{Problem setup}
Following~\cite{epsilonfree}, BayesSim approximates the intractable posterior $p(\bm{\theta}|\mathbf{x}=\mathbf{x}^r)$ by directly learning a conditional density $q_\phi(\bm{\theta}|\mathbf{x})$ parameterised by parameters $\phi$. As we shall see, $q_\phi(\bm{\theta}|\mathbf{x})$ takes the form of a mixture density random feature network. To learn the parameters $\phi$ we first generate a dataset with $N$ pairs $(\bm{\theta}_n,\mathbf{x}_n)$ where $\bm{\theta}_n$ is drawn independently from a distribution $\tilde{p}(\bm{\theta})$ referred to as the \emph{proposal prior}.  
$\mathbf{x}_n$ is obtained by running the simulator with parameter $\bm{\theta}_n$ such that $\mathbf{x}_n = g(\bm{\theta}_n)$. In~\cite{epsilonfree} the authors show that $q_\phi(\bm{\theta}|\mathbf{x})$ is proportional to $\frac{\tilde{p}(\bm{\theta})}{p(\bm{\theta})}p(\bm{\theta}|\mathbf{x})$
when the likelihood $\prod_n q_\phi(\bm{\theta}_n|\mathbf{x}_n)$ is maximised w.r.t. $\phi$. We follow a similar procedure and maximise the log likelihood,
\begin{equation}
{\cal L}(\phi)=\frac{1}{N}\sum_n\log q_\phi(\bm{\theta_n}|\mathbf{x}_n)    
\end{equation}
to determine $\phi$. After this is done, an estimate of the posterior is obtained by

\begin{equation}
    \hat{p}(\bm{\theta}|\mathbf{x}=\mathbf{x}^r)\propto \frac{p(\bm{\theta})}{\tilde{p}(\bm{\theta})}q_\phi(\bm{\theta}|\mathbf{x}=\mathbf{x}^r),
    \label{eq:post}    
\end{equation}
where $p(\bm{\theta})$ is the desirable prior that might be different than the proposal prior. In the case when $\tilde{p}(\bm{\theta})=p(\bm{\theta})$, it follows that $\hat{p}(\bm{\theta}|\mathbf{x}=\mathbf{x}^r)\propto q_\phi(\bm{\theta}|\mathbf{x}=\mathbf{x}^r)$. When $\tilde{p}(\bm{\theta})\neq p(\bm{\theta})$ we need to adjust the posterior as detailed in Section \ref{sec:postrec}.

\subsection{Mixture density random feature networks}
We model the conditional density $q_\phi(\bm{\theta}|\mathbf{x})$ as a mixture of $K$ Gaussians,
\begin{equation}
    q_\phi(\bm{\theta}|\mathbf{x})=\sum_k\alpha_k{\cal N}(\bm{\theta}|\bm{\mu}_k,\bm{\Sigma}_k),
\end{equation}
where $\bm{\alpha}=(\alpha_1,\ldots,\alpha_K)$ are mixing coefficients, $\{\mu_k\}$ are means and $\{\Sigma_k\}$ are covariance matrices. This is analogous to mixture density networks~\cite{bishopmdn} except that we replace the feedforward neural network with Quasi Monte Carlo (QMC) random Fourier features when computing $\bm{\alpha}$, $\bm{\mu}$ and $\bm{\Sigma}$. We justify and describe these features in the next section. 

Denoting $\Phi(\bm{x})$ as the feature vector, the mixing coeficients are calculated as
\begin{equation}
    \bm{\alpha}=\text{softmax}(\mathbf{W}_{\bm{\alpha}} \Phi(\mathbf{x})+\mathbf{b}_{\bm{\alpha}}).
\end{equation}
Note that the operator $\text{softmax}(\mathbf{z}_i)=\frac{\exp(z_i)}{\sum_{k=1}^K \exp{z_k}}$ for $i=1,\ldots,K$ enforces that the sum of coeficients equals to $1$ and each coefficient is between $0$ and $1$.

The means are defined as linear combinations of feature vectors. For each component of the mixture,
\begin{equation}
    \bm{\mu}_k=\mathbf{W}_{\bm{\mu}_k}\Phi(\mathbf{x})+\mathbf{b}_{\bm{\mu}_k}.
\end{equation}

Finally we parametrize the covariance matrices as diagonals matrices with
\begin{equation}
    \text{diag}(\bm{\Sigma}_k)=\text{mELU}(\mathbf{W}_{\bm{\Sigma}_k}\Phi(\mathbf{x})+\mathbf{b}_{\bm{\Sigma}_k})
\end{equation}
where $\text{mELU}$ is a modified exponential linear unit defined as 
\begin{equation}
    mELU(z)= {\begin{cases}\alpha (e^{z}-1)+1 &{\text{for }}z\leq 0\\
    z + 1 &{\text{for }}z>0\end{cases}}
\end{equation}
to enforce positive values. Experimentally this parametrization provided slightly better results than with the exponential function. The diagonal parametrization assumes independence between the dimensions of the simulator parameters $\bm{\theta}$. This turns out to be not too restrictive if the number of components in the mixture is large enough. 

The full set of parameters for the mixture density network is then,
\begin{equation}
    \phi=(\mathbf{W}_{\bm{\alpha}},\mathbf{b}_{\bm{\alpha}}, \{\mathbf{W}_{\bm{\mu}_k},\mathbf{b}_{\bm{\mu}_k},\mathbf{W}_{\bm{\Sigma}_k},\mathbf{b}_{\bm{\Sigma}_k}\}_{k=1}^{K}).
\end{equation}

\subsection{Neural Network features}
BayesSim can use neural network features creating a model similar to the mixture density network in~\cite{bishopmdn}. For a feedforward neural network with two fully connected layers, the features take the form 
\begin{equation}
    \Phi(\mathbf{x})=\sigma(\mathbf{W}_2(\sigma(\mathbf{W}_1\mathbf{x}+\mathbf{b}_1))+\mathbf{b}_2),
\end{equation}
where $\sigma(\cdot)$ is a sigmoid function; we use $\sigma(\cdot)=\tanh(\cdot)$ in our experiments.  
This network structure was used in the experiments and compared to the Quasi Monte Carlo random features described below.

\subsection{Quasi Monte Carlo random features}
BayesSim can use random Fourier features~\cite{rahimi08rff} instead of neural nets to parameterise the mixture density. There are several reasons why this can be good choice. Notably, 1) random Fourier features -- of which QMC features are a particular type -- approximate possibly infinite Hilbert spaces with properties defined by the choice of the associated kernel. In this way prior information about properties of the function space can be readily incorporated by selecting a suitable positive semi-definite kernel; 2) the approximation converges to the original Hilbert space with order ${\cal O}(1/\sqrt{s})$, where $s$ is the number of features, therefore independent of the input dimensionality;  3) experimentally, we verified that mixture densities with random Fourier features are more stable to different initialisations and converge to the same local maximum in most cases. 

Random Fourier features approximate a shift invariant kernel $k(\bm\tau)$, where $\bm\tau=\|\mathbf{x} - \mathbf{x}'\|$, by a dot product $k(\bm\tau)\approx \Phi(\mathbf{x})^T\Phi(\mathbf{x})$ of finite dimensional features $\Phi(\mathbf{x})$.  
This is possible by first applying the Bochner's theorem~\cite{stein1999spatialdata} stated below: \\

\noindent{\bf Theorem 1} {\em (Bochner's Theorem) A shift invariant 
kernel $k(\bm\tau)$, $\bm\tau\in\mathbb{R}^D$, associated with a positive
finite measure $d\mu\left(\bm{\omega}\right)$ can be represented in terms of its
Fourier transform as,
\begin{equation}
k(\bm\tau)=\int_{\mathbb{R}^D}e^{- i \bm{\omega}\cdot \bm{\tau}} d\mu\left(\bm{\omega}\right).
\label{eq:bochner}
\end{equation}}
\noindent The proof can be found in~\cite{Gihman74}. When $\mu$ has density ${\cal K}(\bm\omega)$ then ${\cal K}$ represents the spectral distribution for a positive semi-definite $k$. In this case $k(\bm\tau)$ and ${\cal K}(\bm\omega)$ are Fourier duals:
\begin{equation}
    k(\bm\tau)=\int{\cal K}(\bm\omega)e^{- i\bm\omega\cdot\bm\tau}d\bm\omega.
    \label{eq:duals}
\end{equation}

Approximating Equation \ref{eq:duals} with a Monte Carlo estimate with $N$ samples, yields
\begin{equation}
    k(\bm\tau)\approx \frac{1}{N} \sum_{n=1}^N (e^{-i \bm{\omega}_n \mathbf{x}})(e^{-i \bm{\omega}_n\mathbf{x}'}), 
\end{equation}
where $\bm\omega$ is sampled from the density ${\cal K}(\bm\omega)$. 

 Finally, using Euler's formula ($e^{-ix}=\cos(x) - i\sin(x)$) we recover the features:

\begin{equation}
\begin{split}
\Phi(\mathbf{x})=\frac{1}{\sqrt{N}}[\cos\left(\bm{\omega}_1\mathbf{x}+b_1\right),
\ldots,\cos\left(\bm{\omega}_n\mathbf{x}+b_n\right), \\
-i\cdot\sin\left(\bm{\omega}_1\mathbf{x}+b_1\right),\ldots,-i\cdot\sin\left(\bm{\omega}_n\mathbf{x}+b_n\right)].
\label{eq:rbf_fourier}
\end{split}
\end{equation}
where bias terms $\mathbf{b}_i$ are introduced with the goal of rotating the projection and allowing for more flexibility in capturing the correct frequencies.  

This approximation can be used with all shift invariant kernels proving flexibility in introducing prior knowledge by selecting a suitable kernel for the problem. For example, the RBF kernel can be approximated using the features above with $\bm\omega\sim {\cal N}(0,2\sigma^{-2}I)$ and $b \sim {\cal U}[-\pi,\pi]$. $\sigma$ is a hyperparameter that corresponds to the kernel length scale and is usually set up with cross validation. 

We further adopt a quasi Monte Carlo strategy for sampling the frequencies. In particular we use Halton sequences~\cite{braaten1979halton} which has been shown in 
~\cite{avron16qmc} to have better convergence rate and lower approximation error than standard Monte Carlo. 

\subsection{Posterior recovery}
\label{sec:postrec}
From Equation \ref{eq:post} we note that when the proposal prior is different than the desirable prior, we need to adjust the posterior by weighting it with the ratio $p(\bm{\theta})/\tilde{p}(\bm{\theta})$. 

In this paper we assume the prior to be uniform, either with finite support -- defined within a range and zero elsewhere -- or improper, constant value everywhere. Therefore,
\begin{equation}
    \hat{p}(\bm{\theta}|\mathbf{x}=\mathbf{x}^r)\propto\frac{q_\phi(\bm{\theta}|\mathbf{x}^r)}{\tilde{p}(\bm{\theta})}.
\end{equation}

When the proposal prior is Gaussian, we can compute the division between a mixture and a single Gaussian analytically. In this case, since $q_\phi(\bm{\theta}|\mathbf{x})$ is a mixture of Gaussians and $\tilde{p}(\bm{\theta}) \sim {\cal N}(\bm\theta|\bm\mu_0,\bm\Sigma_0)$, the solution is given by 
\begin{equation}
    \hat{p}(\bm{\theta}|\mathbf{x}=\mathbf{x}^r) = \sum_k\alpha_k'{\cal N}(\bm{\theta}|\bm{\mu}_k',\bm{\Sigma}_k')
\end{equation}
where,
\begin{align}
    \bm{\Sigma}_k' &= \left(\bm{\Sigma}_k^{-1}-\bm{\Sigma}_0^{-1}   \right)^{-1}\\
    \bm{\mu}_k' &= \bm{\Sigma}_k^{-1}\left(\bm{\Sigma}_k^{-1}\bm{\mu}_k - \bm{\Sigma}_0^{-1}\bm{\mu}_0 \right)\\
    \alpha_k' &= \frac{\alpha_k\exp(-\frac{1}{2}\lambda_k)}{\sum_{k'}\alpha_{k'}\exp(-\frac{1}{2}\lambda_{k'})},
\end{align}
and the coefficients ${\lambda_k}$ are given by
\begin{multline}
    \lambda_k=\log\det\bm{\Sigma}_k-\log\det\bm\Sigma_0-\log\det\bm\Sigma_k'+\bm\mu_k^T\bm\Sigma_k^{-1}\bm\mu_k \\
    -\bm\mu_0^T\bm\Sigma_0^{-1}\bm\mu_0 - \bm\mu_k'^T\bm\Sigma_k'^{-1}\bm\mu_k'.
\end{multline}

\subsection{Sufficient statistics for state-action trajectories}
\label{sec:ss}
Trajectories of state and action pairs in typical problems can be long sequences making the input dimensionality to the model prohibitive large and computationally expensive. We adopt a strategy commonly used in ABC; instead of inputting raw state and action sequences to the model, we first compute some sufficient statistics. Formally, $\mathbf{x}=\psi(\mathbf{S}, \mathbf{A})$ where  $\mathbf{S}=\{\mathbf{s}^t\}_{t=1}^T$ and $\mathbf{A}=\{\mathbf{a}^t\}_{t=1}^T$ are sequences of states and actions from $t=1$ to $T$. There are many options in the literature for sufficient statistics for time series or trajectory data. For example, the mean, log variance and autocorrelation for each time series as well as cross-correlation between two time series. Another possibility is to learn these from data, for example with an unsupervised encoder-decoder recurrent neural network~\cite{srivastava15}. However, such a representation would need to be trained with simulated trajectories and might not be able to capture complexities in the real trajectories. This will be investigated in future work. Here we adopt a simpler strategy and use statistics commonly applied to stochastic dynamic systems such as the Lotka-Volterra model~\cite{wilkinson2011}. 

Defining $\bm{\tau}=\{\mathbf{s}^t-\mathbf{s}^{t-1}\}_{t=1}^T$ as the difference between immediate future states and current states, the statistics 
\begin{equation}
    \psi(\mathbf{S},\mathbf{A})=(\{\langle \bm{\tau}_i, \mathbf{A}_j \rangle\}_{i=1,j=1}^{D_s,D_a}, \mathrm{E}[\bm{\tau}],\mathrm{Var}[\bm{\tau}]),
\end{equation}
where $D_s$ is the dimensionality of the state space, $D_a$ is the dimensionality of the action space, $\langle \cdot,\cdot \rangle$ denotes the dot product, $\mathrm{E}[\cdot]$ is the expectation, and $\mathrm{Var}[\cdot]$ the variance.
    
\subsection{Example: CartPole posterior}

\begin{figure*}[t]
    \centering
    \includegraphics[width=0.325\textwidth]{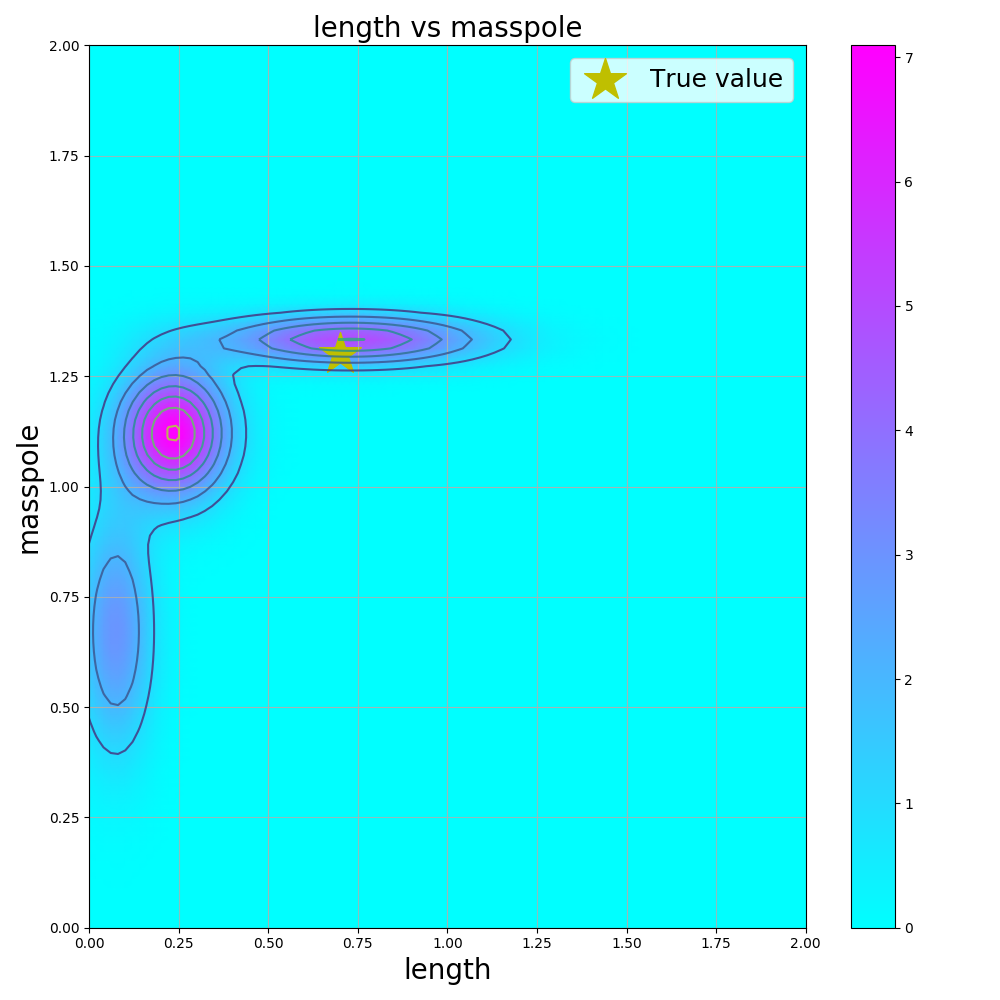}
    \includegraphics[width=0.325\textwidth]{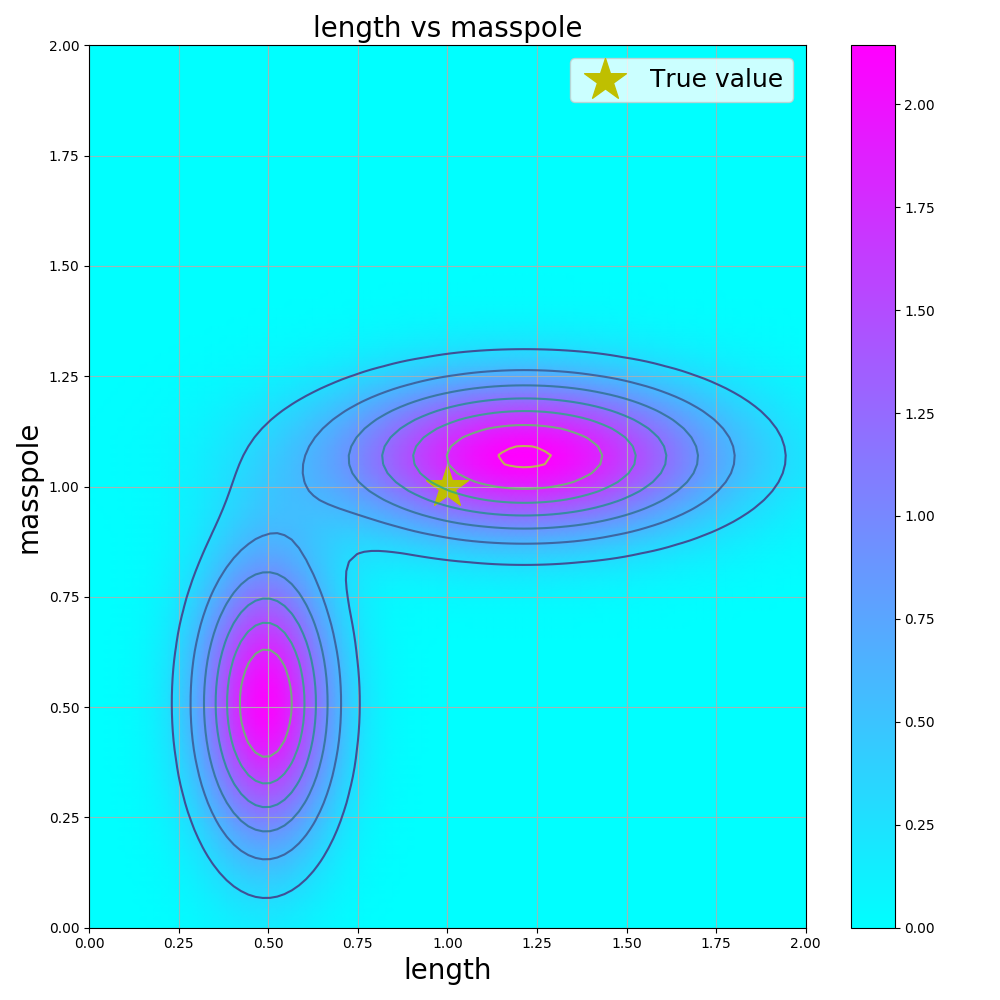}
    \includegraphics[width=0.325\textwidth]{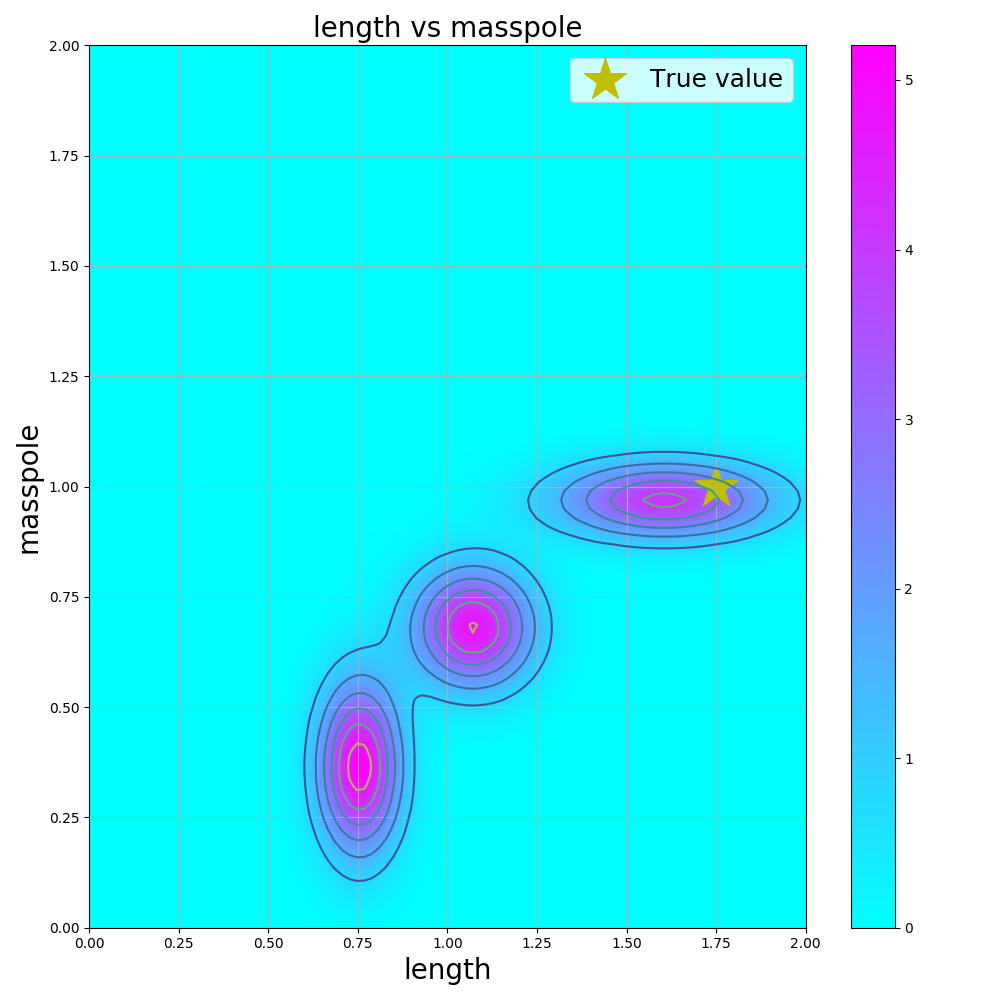}
    \caption{Example of joint posteriors obtained for the CartPole problem with different parametrizations for \texttt{length} and \texttt{masspole}. The true value is indicated by a star. Note that the joint posteriors capture the multimodality of the problem when two or more explanations seem likely, for example, a longer pole \texttt{length} with a lighter \texttt{masspole} or vice versa.}
    \label{fig:cartpole}
\end{figure*}

We provide a simple example to demonstrate the algorithm in estimating unknown simulation parameters for the famous CartPole problem. In this problem a pole installed on a cart needs to be balanced by applying forces to the left or to the right of the cart. For this example we assume that both the mass and the length of the pole are not available and we use BayesSim to obtain the posterior for these parameters. We assume uniform priors for both parameters and collect 1000 simulations following a rl-zoo policy~\footnote{https://github.com/araffin/rl-baselines-zoo} to train BayesSim. With the model trained, we collected 10 trajectories with the correct parameters to simulate the real observations. Figure \ref{fig:cartpole} shows the posteriors for both problems. As with many problems involving two related variables, \texttt{masspole} and pole \texttt{length} exhibit statistical dependencies that generate multiple explanations for their values. For example, the pole might have lower mass and longer length, or vice versa. BayesSim is able to recover the multi-modality nature of the posterior providing densities that represent the uncertainty of the problem accurately.

\subsection{Domain randomization with BayesSim}
Here we describe the domain randomization strategy to take full advantage of the posterior obtained by the inference method. 
Given the posterior obtained from the simulation parameters $\hat{p}(\bm\theta|\mathbf{x}=\mathbf{x}^r)$ we maximize the objective,
\begin{equation}
  J(\bm\beta) = \mathbb{E}_{\bm\theta} \left[\mathbb{E}_{\bm\eta}\left[ \sum_{t-0}^{T-1} \gamma^{(t)}r(\mathbf{s}_t, \mathbf{a}_t) | \bm\beta \right]\right], 
  \label{eq:domain_cost}
 \end{equation}
where $\bm\theta \sim \hat{p}(\bm\theta|\mathbf{x}=\mathbf{x}^r) $
with respect to the policy parameters $\bm\beta$. Since the posterior is a mixture of Gaussians, the first expectation can be approximated by sampling a mixture component following the distribution over $\bm\alpha$ to obtain a component $k$, followed by sampling the corresponding Gaussian ${\cal N}(\bm\theta|\bm\mu_k,\bm\Sigma_k)$.

\section{Experiments}
\label{sec:experiments}

Experiments are presented in two different cases to demonstrate and assess the performance of BayesSim. In Section \ref{sec:exp_post} we verify and compare the accuracy of the posterior recovered. In Section \ref{sec:exp_robust} we compare the robustness of policies trained by randomizing following the prior versus posterior distribution over simulation parameters.    

\subsection{Posterior recovery}
\label{sec:exp_post}
The first analysis we carry out is the quality of the posteriors obtained for different problems and methods. We use the log probability of the target under the mixture model as the measure, defined as $\log p(\bm\theta_*|\mathbf{x}=\mathbf{x}^r)$, where $\bm\theta_*$ is the actual value for the parameter. We compare Rejection-ABC~\cite{pritchard1999population} as the baseline, the recent $\epsilon$-Free~\cite{epsilonfree} which also provides a mixture model as the posterior, and BayesSim using either a two layer neural network with 24 units in each layer, and BayesSim with quasi random Fourier Features. For the later we use the Matern 5/2 kernel~\cite{rasmussen2005gp} and set up the the sampling precision $\sigma$ by cross validation. Three different simulators were used for different problems; OpenAI Gym~\cite{gym2016}, PyBullet~\footnote{https://pypi.org/project/pybullet/}, and MuJoCo~\cite{todorov2012mujoco}. Finally, the following problems were considered; 
CartPole (Gym), Pendulum (Gym), Mountain Car (Gym), Acrobot (Gym), Hopper (PyBullet), Fetch Push (MuJoCo) and Fetch Slide (MuJoCo). For all configurations of methods and parameters, training and testing were performed 5 times with the log probabilities averaged and standard deviation computed. To extract the real observations, we simulate the environments with the actual parameters 10 times and average the sufficient statistics to obtain $\mathbf{x}^r$. In all cases we collect sufficient statistics by performing rollouts for either a maximum of 200 time steps or until the end of the episode.

Table \ref{tab:logprob} shows the results (means and standard deviations) for the log probabilities. BayesSim with either RFF or Neural Network features provides generally higher log-probabilities and lower standard deviation than Rejection ABC. This indicates that the posteriors provided by BayesSim are more peaked and centered around the correct values for the parameters. Compared to $\epsilon$-Free, the results are equivalent in terms of the means but BayesSim generally provides lower standard deviation across multiple runs of the method, indicating it is more stable than $\epsilon$-Free. Comparing BayesSim with RFF and NN, the RFF features lead to higher log probabilities in most cases but BayesSim with neural networks have lower standard deviation. 

\begin{figure}
    \centering
    \includegraphics[width=0.45\textwidth]{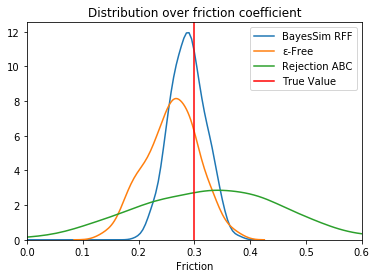}
    \caption{Posteriors recovered by different methods for the Fetch slide problem. Note that BayesSim with random features provides a posterior that is more peaked around the true value.}
    \label{fig:friction_post}
\end{figure}

These results suggest that BayesSim with either RFF or NN is comparable to the state-of-art, and in many cases superior when estimating the posterior distribution over the simulation parameters. For the robotics problems analyzed in the next section, however, BayesSim with RFF provide significant superior results than the other methods and slightly better than BayesSim with NN. This can be better observed when we plot the posteriors in Figure \ref{fig:friction_post}. BayesSim RFF is significantly more peaked and centered around the true friction value.

\begin{table*}[t]
\centering
\begin{tabular}{|l|l|c|c|c|c|c|}
    \hline
    Problem & Parameter & Uniform prior & Rejection ABC & $\epsilon$-Free & BayesSim RFF & BayesSim NN  \\
    \hline
    CartPole & pole length & $[0.1, 2.0]$ & -0.342$\pm$0.15 & \bf{-0.211$\pm$0.07} & -0.609$\pm$0.39 & -0.657$\pm$0.25 \\
             & pole mass & $[0.1, 2.0]$ & 0.032$\pm$0.21 & 0.056$\pm$0.14 & \bf{0.973 $\pm$ 0.26} & 0.633$\pm$ 0.52 \\
    \hline         
    Pendulum & dt & $[0.01, 0.3]$ & 2.101$\pm$1.04 & 2.307$\pm$0.84  &3.192$\pm$0.30 & \bf{3.199$\pm$0.17}\\
    \hline
    Mountain Car & power & $[0.0005, 0.1]$ & 3.69$\pm$1.21 & 3.800$\pm$1.06 & 3.863$\pm$0.52 & \bf{3.901$\pm$0.2}  \\
    \hline
    Acrobot & link mass 1 & $[0.5, 2.0]$ & 1.704$\pm$0.82 & 1.883$\pm$0.79 & \bf{2.046$\pm$0.37} & 1.331$\pm$0.22 \\
            & link mass 2 & $[0.5, 2.0]$ & 1.832$\pm$0.93 & \bf{2.237$\pm$0.76} & 0.321$\pm$1.85 & 1.513$\pm$0.39 \\
            & link length 1 & $[0.1, 1.5]$ & \bf{2.421$\pm$0.75} & 2.135$\pm$0.50 & 2.072$\pm$0.76 & 1.856$\pm$0.18 \\
            & link length 2 & $[0.5, 1.5]$ & -0.521$\pm$0.36 & -0.703$\pm$0.16 & \bf{-0.148$\pm$0.19} & -0.672$\pm$0.09 \\
    \hline
    Hopper & lateral friction & $[0.3, 0.5]$ & 3.032$\pm$0.43 &  3.154$\pm$0.81 & 2.622$\pm$0.64 & \bf{3.391$\pm$0.08}\\
    \hline
    Fetch Push & friction & $[0.1, 1.0]$ & 1.332$\pm$0.54 & 2.013$\pm$0.09 & \bf{2.423$\pm$0.07} &  2.404$\pm$0.05  \\
    \hline
    Fetch Slide & friction & $[0.1, 1.0]$ & 1.014$\pm$0.38 & 1.614$\pm$0.12 & \bf{2.391$\pm$0.06} & 2.111$\pm$0.03  \\
    \hline
\end{tabular}
\caption{Mean and standard deviation of log predicted probabilities for several likelihood-free methods, applied to seven different problems and parameters.}
\label{tab:logprob}
\end{table*}

\subsection{Robustness of policies}
\label{sec:exp_robust}
We evaluate robustness of policies by comparing their performance on the uniform prior and the learned posterior provided by BayesSim. Evaluation is done over a pre-defined range of simulator settings and the average reward is shown for each parameter value.

\begin{figure*}[t]
    \centering
    \includegraphics[width=0.9\columnwidth]{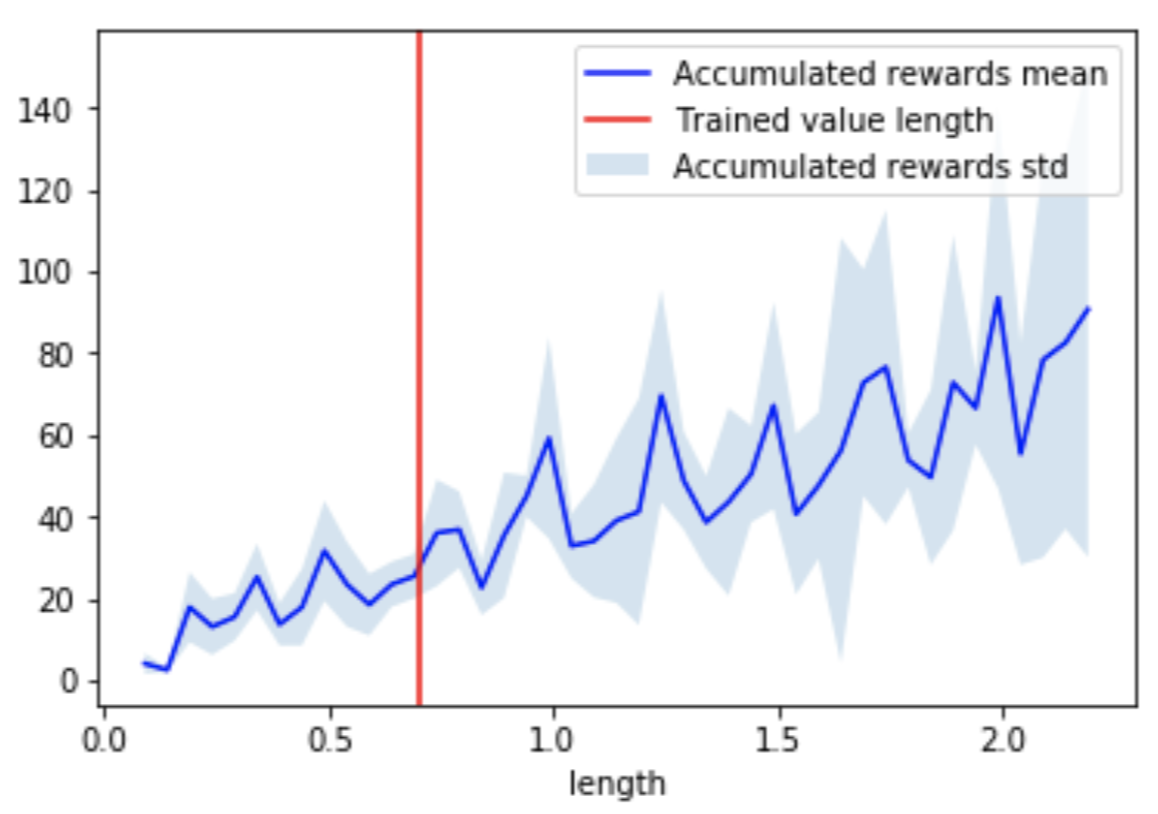}
    \includegraphics[width=0.9\columnwidth]{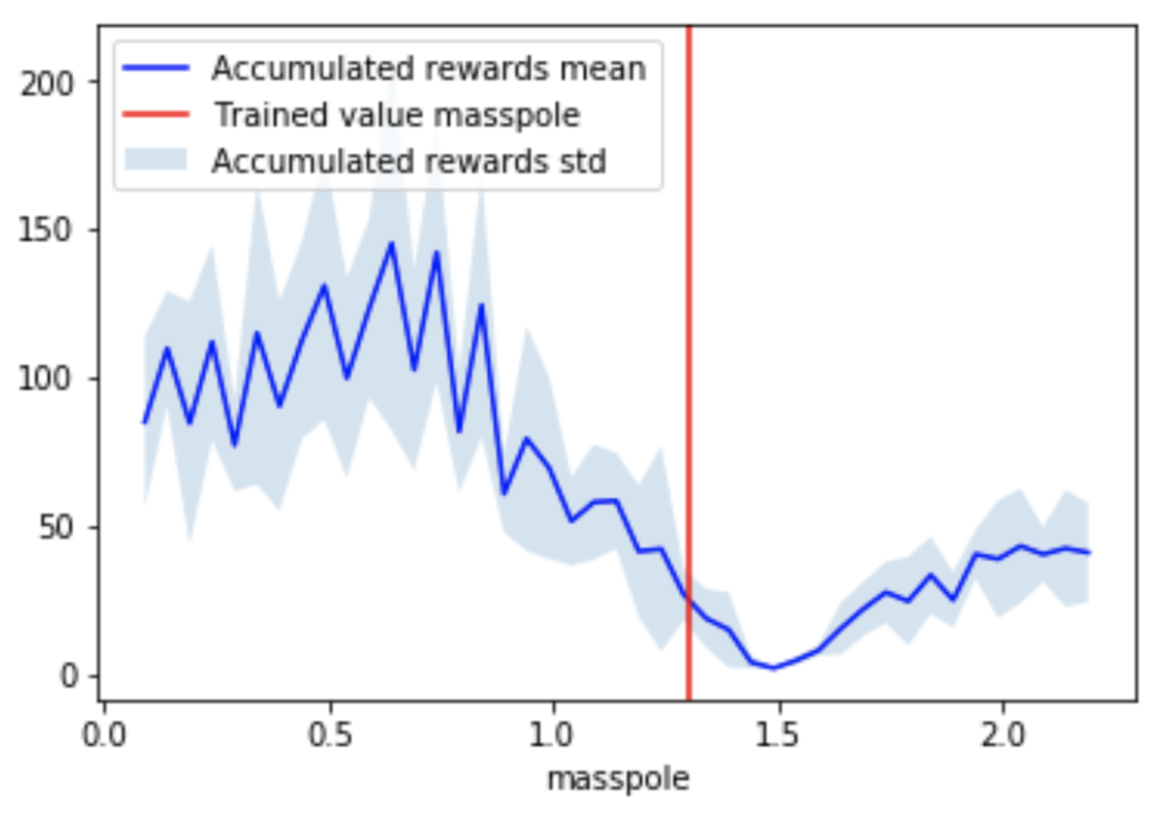}\\
    \includegraphics[width=0.9\columnwidth]{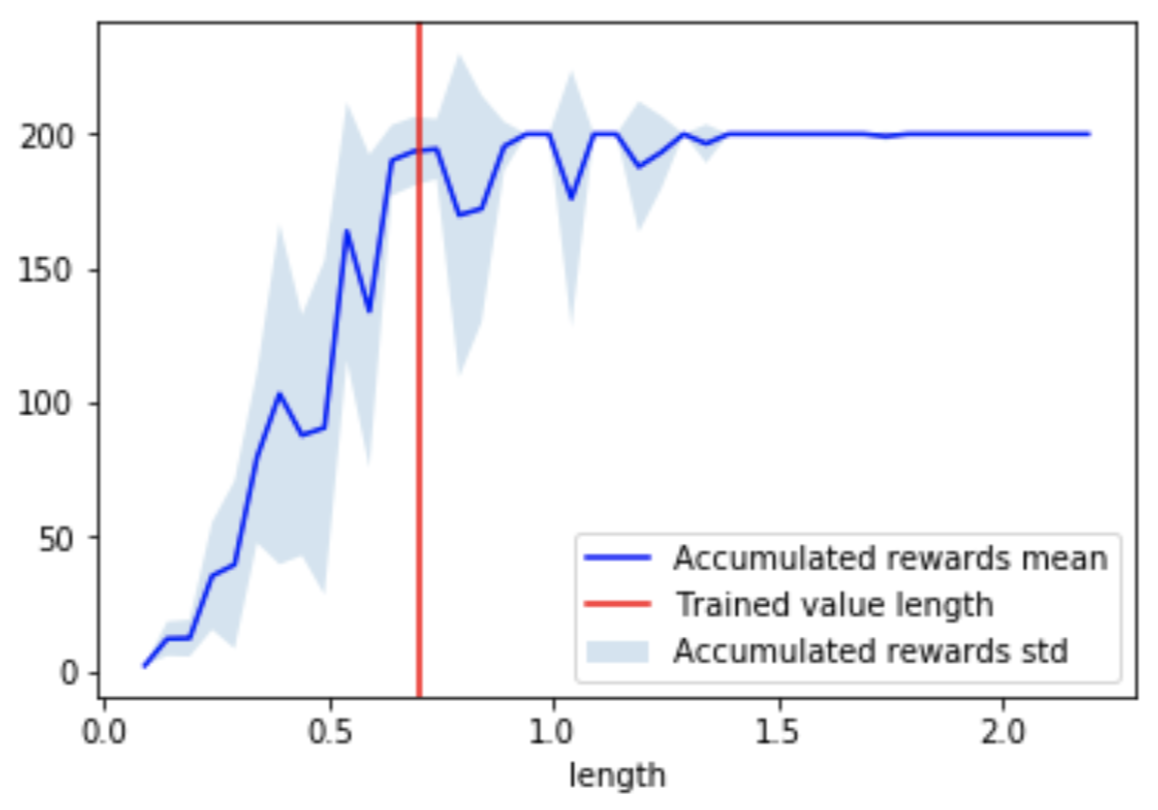} 
    \includegraphics[width=0.9\columnwidth]{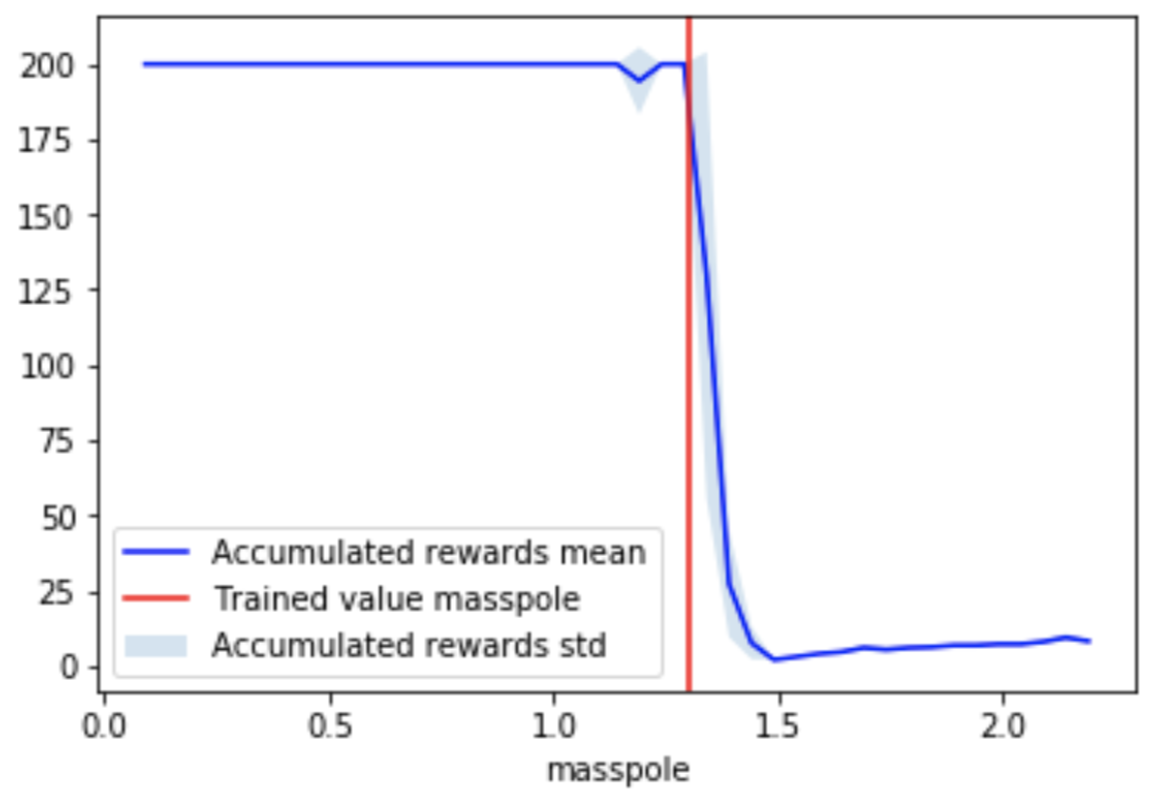}
    \caption{Accumulated rewards for CartPole policies trained with PPO by randomizing over prior and posterior joint densities. Top left: Performance of the policy trained with the prior, over parameter \texttt{length}. \texttt{masspole} is set to actual. Top right: Similar to top left, but over multiple \texttt{masspole} values. Bottom left: Performance of policy trained with the posterior, over parameter \texttt{length}. Bottom right:  Similar to bottom left, but over multiple \texttt{masspole} values.}    
    \label{fig:robustness_cartpole}
\end{figure*}

In the first set of experiments we use the CartPole problem as a simple example to illustrate the benefits of posterior randomization. We trained two policies, the first randomizing with a uniform prior for {\em length} and {\em masspole} as indicated in Table~\ref{tab:logprob}. The second, randomized based on the posterior provided by BayesSim with RFF. In both cases we use PPO to train the policies with 100 samples from the prior and posterior, for 2M timesteps. The results are presented in Figure~\ref{fig:robustness_cartpole}, averaged over several runs with the corresponding standard deviations. It can be observed that randomization over the posterior yields a significantly more robust policy, in particular at the actual parameter value. Also noticeable is the reduction in performance for lower {\em length} values and higher {\em masspole} values. This is expected as it is more difficult to control the pole position when the length is short due to the increased dynamics of the system. Similarly, when the mass increases too much, beyond the value it was actually trained on, the controller struggles to maintain the pole balanced. Importantly, the policy learned with the posterior seems much more stable across multiple runs as indicated by the lower variance in the plots. 

In the second set of experiments we use a Fetch robot available in OpenAI Gym \cite{brockman2016openai} to perform both push and slide tasks. The first is a closed loop scenario, where the arm is always in range of the entire table and, hence,  it can correct its trajectories according to the input it receives from the environment. The second is a more difficult open loop scenario, where the robot has usually only one shot at pushing the puck to its desired target. For both tasks, the friction coefficient of the object and the surface plays a major role in the final result as they are strictly related to how far the object goes after each force is applied. A very low friction coefficient means that the object is harder to control as it slides more easily and a very high one means that more force needs to be applied in order to make the object to move.

Our goal is to recover a good approximation of the posterior over friction coefficients using BayesSim. Initially, we need to learn a policy with a fixed friction coefficient that will be used for data generation purposes. We train this policy using DDPG with experiences being sampled using HER for 200 epochs with 100 episodes/rollouts per epoch. Gradient updates are done using Adam with step size of 0.001. We then run this policy multiple times with different friction coefficients in order to approximate the likelihood function and recover the full posterior over simulation parameters. With the dynamics model in hand, we can finally recover the desired posterior using some data sampled from the environment we want to learn the dynamics from. Training is carried out using the same aforementioned settings but instead of using a fixed friction coefficient, we sample a new one from its respective distribution every time a new episode starts.

\begin{figure}
    \centering
    \includegraphics[width=.95\columnwidth]{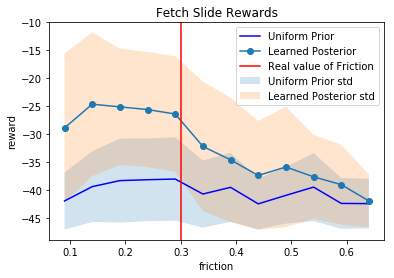} \\
    \includegraphics[width=.95\columnwidth]{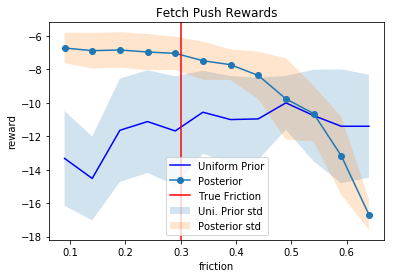}
    \caption{Comparison between policies trained on randomizing the prior vs BayesSim posterior for different values of the simulation parameter. Top: Fetch slide problem. Bottom: Fetch push problem.}    
    \label{fig:robustness_fetch}
\end{figure}

The results from both tasks are presented on Figure \ref{fig:robustness_fetch}. As it has been shown in previously work \cite{robcontrol}, the uniform prior works remarkably well on the push task. This happens as the robot has the opportunity to correct its trajectory whether something goes wrong. As it has been exposed to a wide range of scenarios involving different dynamics, it can then use the input of the environment to perform corrective actions and still be able to achieve its goal. However, the results for slide task differ significantly since using a wide uniform prior has led the robot to achieve a very poor performance. This happens as not only the actions for different coefficients in most times are completely different but also because the robot has no option of correcting its trajectory. This is where methods like BayesSim are useful as it recovers a distribution with very high density around the true parameter and, hence, leads to a better overall control policy. Our results shows that higher rewards are achieved around the true friction value while the uniform prior results are mostly flat throughout all values.

\section{Conclusions} 
\label{sec:conclusion}

This paper represents the first step towards a Bayesian treatment of robotics simulation parameters, combined with domain randomization for policy search. Our approach is connected to system identification in that both attempt to estimate dynamic models, but ours uses a black-box generative model, or simulator, totally integrated into the framework. Prior distributions can also be provided and incorporated into the model to compute a full, potentially multi-modal posterior over the parameters. The method proposed here, BayesSim, performs comparably to other state-of-the-art likelihood-free approaches for Bayesian inference but appears more stable to different initializations, and across multiple runs when recovering the true posterior. Finally, we show that domain randomization with the posterior leads to more robust policies over multiple parameter values compared to policies trained on uniform prior randomization. 

The two applications described in the paper for likelihood-free inference are two instances of a large range of problems where simulators can make use of a full set of parametrizations to best represent reality. In this manner, our framework can be integrated in many other problems involving simulators. An interesting line of research for future work is to use BayesSim to help simulators synthesize images by randomizing over background properties. This can potentially help in making many computer vision problems more robust to environment variability in many tasks including object recognition, 3D pose estimation, or motion tracking.    

As typical in the likelihood-free inference literature, BayesSim relies on the definition of meaningful sufficient statistics for the trajectories of states and actions. Alternatively, a lower dimensional representation for the trajectories could be created using recent encoder-decoder methods and recurrent neural networks known to perform well for time series prediction such as LSTMs~\cite{Hochreiter1997}. Hence, the entire framework can be learnt end to end. This is an interesting area for future development but careful consideration should be given to potential overfitting to simulation data. LSTMs usually require a lot of data for training therefore most of the training trajectories will be generated from simulated trajectories. This can introduce undesirable specific characteristics of the simulator in the low dimensional representation that are not observed in real trajectories, making the representation less sensitive to variations in the simulator parameters to be estimated. Despite this, automating the process of generating robust statistics from trajectories remains a valuable direction for future research.           

\bibliographystyle{plainnat}
\bibliography{references}
\end{document}